\documentclass[10pt,twocolumn,letterpaper]{article}

\usepackage[pagenumbers]{cvpr} 

\usepackage{graphicx}
\usepackage{amsmath}
\usepackage{amssymb}
\usepackage{booktabs}

\usepackage[pagebackref,breaklinks,colorlinks]{hyperref}

\usepackage[capitalize]{cleveref}
\crefname{section}{Sec.}{Secs.}
\Crefname{section}{Section}{Sections}
\Crefname{table}{Table}{Tables}
\crefname{table}{Tab.}{Tabs.}

\def\cvprPaperID{*****} 
\def\confName{CVPR}
\def\confYear{2023}

\begin{document}

\title{3rd Place Solution for PVUW2023 VSS Track: A Large Model for Semantic Segmentation on VSPW}

\author{
Shijie Chang$^1$\protect\footnotemark[2] ~
Zeqi Hao$^1$\protect\footnotemark[2] ~
Ben Kang$^1$ ~
Xiaoqi Zhao$^1$ ~
Jiawen Zhu$^1$ ~
Zhenyu Chen$^1$ ~\\
Lihe Zhang$^1$ ~
Lu Zhang$^1$ ~
Huchuan Lu$^{1,2}$\\ 
$^1$School of Information and Communication Engineering, Dalian University of Technology \\ 
$^2$Peng Cheng Laboratory \\
\protect\footnotemark[2]  equal contribution \\
{\tt\small \{csj, hzq, kangben, zxq, jiawen, dlutczy\}@mail.dlut.edu.cn,}\\ {\tt\small \{zhanglihe, zhangluu,  lhchuan\}@dlut.edu.cn}
}
\maketitle

\begin{abstract}
   In this paper, we introduce 3rd place solution for PVUW2023 VSS track. Semantic segmentation is a fundamental task in computer vision with numerous real-world applications. We have explored various image-level visual backbones and segmentation heads to tackle the problem of video semantic segmentation. Through our experimentation, we find that InternImage-H as the backbone and Mask2former as the segmentation head achieves the best performance. In addition, we explore two post-precessing methods: CascadePSP and Segment Anything Model (SAM). Ultimately, our approach obtains 62.60\% and 64.84\% mIoU on the VSPW test set1 and final test set, respectively, securing the third position in the PVUW2023 VSS track.
\end{abstract}

\section{Introduction}
\label{sec:intro}

Pixel-level Scene Understanding is one of the fundamental problems in computer vision, which aims at recognizing object classes, masks and semantics of each pixel in the given image. The 2nd Pixel-level Video Understanding in the Wild (PVUW2023) consists of two challenge tracks, VSS and VPS, which respectively use the VSPW \cite{miao2021vspw} and VIPSeg \cite{miao2022large} datasets as the test sets. We selects the VSS track, which aims to perform pixel-level predictions of pre-defined semantic categories on the VSPW dataset.

In recent years, unified macromodels, also known as large-scale models, have emerged as powerful tools in the field of computer vision. These models have demonstrated exceptional performance in fundamental tasks such as image classification, object detection, and image segmentation. The key advantage of unified macromodels lies in their powerful feature modeling capability, allowing them to learn rich and abstract representations from data. By leveraging their large number of parameters and complex network architectures, unified macromodels can capture intricate patterns and high-level semantic information from raw input data. The learned representations can be generalized effectively to downstream tasks, facilitating knowledge transfer and reducing the need for task-specific model architectures. Moreover, deep learning heavily relies on the availability of specialized data for training. The abundance of data plays a vital role in enabling deep models to learn complex representations and generalize well to new examples. 

Based on the advantages of large models and diverse professional data, we have developed a novel model using Internimgage-H \cite{wang2022internimage, zhu2022uni1, zhu2022uni, li2022uni, yang2022bevformer, su2022towards, li2022bevformer} as the encoder and Mask2former \cite{cheng2021Mask2former} as the decoder. This architecture allows us to take advantage of the powerful feature modeling capabilities of Internimgage-H and the semantic capture abilities of Mask2former. To further enhance the feature modeling and semantic understanding of our model, we have employed a pre-training strategy using the COCO-Stuff \cite{caesar2018cocostuff} dataset. Finally, our method obtained 62.60\% and 64.84\% mIoU on the VSPW test set1 and final test set, respectively, and achieved the third place on the PVUW2023 VSS track.

\section{Method}
\label{sec:method}

In this section, we first present the pipeline of our method. Then, we introduce each component that we have attempted in our approach. Finally, we describe training and inference process of our method.

\begin{figure*}[t]
\begin{center}
\includegraphics[width=1.0\linewidth]{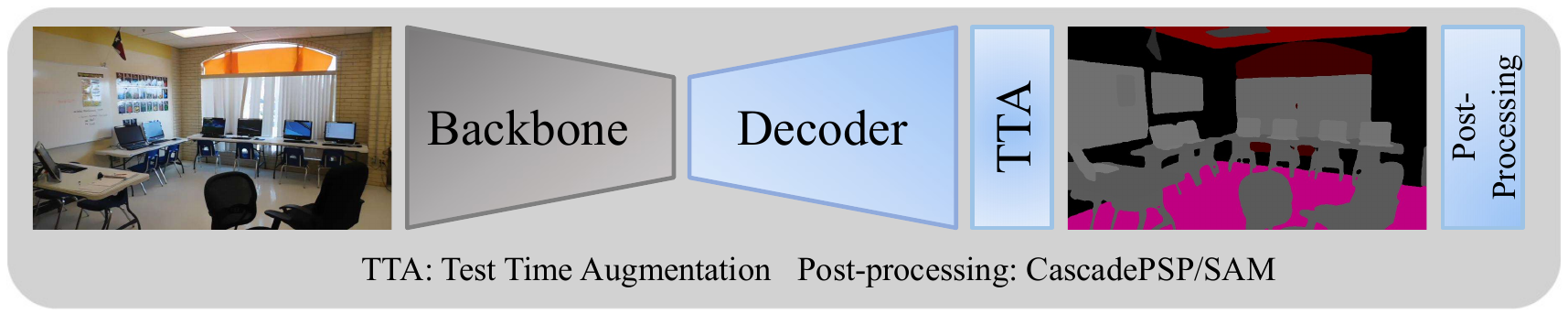}
\end{center}
\vspace{-5mm}
\caption{Overall of our method. It consists of an encoder, a decoder, test time augmentation and a post-processing stage.}
\vspace{-3mm}
\label{fig:fig1}
\end{figure*}

\subsection{Pipeline}

As illustrated in Fig.~\ref{fig:fig1}, we employ an image-level approach to address VSS, which consists of an encoder, a decoder, test time augmentation and a post-processing stage.
Given an input image, our model first utilizes an encoder to extract feature maps. The extracted feature maps are then fed into the decoder to obtain the prediction. Finally, a post-processing method is employed to refine the prediction.
We employ Swin Transformer \cite{liu2021Swin}, ConvNeXt \cite{liu2022convnet} and InternImage \cite{wang2022internimage} as the backbone, UperNet \cite{xiao2018unified} and Mask2former \cite{cheng2021Mask2former} as the segmentation heads. As for post-processing method, we attempt CascadePSP \cite{cheng2020cascadepsp} and SAM \cite{kirillov2023segany}.

\subsection{Encoder}

\textbf{Swin Transformer.} Swin Transformer \cite{liu2021Swin} is a general-purpose backbone for computer vision. Swin Transformer modifies the global self-attention mechanism of Vision Transformer to a shifted-window self-attention, reducing the computational complexity of self-attention. Swin Transformer's impressive  performance in downstream tasks like object detection and semantic segmentation positions it as one of the prevailing backbones. Swin-L achieves 53.5\% mIoU on ADE20K \cite{zhou2019semantic}. Due to its high performance, we employ is as the backbone.

\textbf{ConvNeXt.} ConvNeXt \cite{liu2022convnet} is an open-source model based on convolutional neural networks (CNNs). It extensively conducts ablation study on various techniques of CNNs, while also analyzing and importing techniques from vision transformer models like Swin Transformer. By integrating the strengths of CNNs and the micro design of Swin transformer, it further explores the performance of CNNs and achieves exceptional performance. Unlike Swin Transformer, ConvNeXt is based on CNNs, making it a complementary backbone when compared to Swin Transformer. Therefore, we also utilize ConvNeXt as a backbone to complement Swin Transformer.

\textbf{InternImage.} Internimage \cite{wang2022internimage} is a new large-scale CNN-based foundation model which effectively scales to over 1 billion parameters and 400 million training images and achieves comparable or even better performance than state-of-the-art ViTs. Different from the recent CNNs that focus on large dense kernels, InternImage takes deformable convolution as the core operator, so that it not only has the large effective receptive field required for downstream tasks such as detection and segmentation, but also has the adaptive spatial aggregation conditioned by input and task information. As a result, InternImage reduces the strict inductive bias of traditional CNNs and makes it possible to learn stronger and more robust patterns with large-scale parameters from massive data like ViTs. With its powerful object representation capabilities, Internimgage has demonstrated impressive performance on various representative computer vision tasks. For example, InternImage-H has achieved an improvement of 89.6\% top-1 accuracy on ImageNet \cite{ILSVRC15}, and achieves 62.9\% mIoU and 65.4\% mAP on the challenging downstream benchmarks ADE20K \cite{zhou2019semantic} and COCO \cite{lin2014microsoft}, respectively. Besides, it's worth mentioning that internimgage-H/G is one of the few open-source large models available. We can not only utilize the model structure, but also download pre-trained weights to fine-tune the model on the VSPW dataset. Unfortunately, the best-performing model that composed of InternImage-H and the Mask2former for semantic segmentation are not open-source. As a result, we have to implement it ourselves.

\subsection{Decoder}

\textbf{UperNet.} UperNet \cite{xiao2018unified} is a multi-task model based on Feature Pyramid Network (FPN), which is widely employed in semantic segmentation. We selected UperNet as the segmentation head due to its simplicity and effectiveness, as well as most backbones are typically used in conjunction with UperNet. In addition, we can easily obtain pre-trained weights of UperNet on other segmentation datasets such as ADE20K and Cityscapes \cite{Cordts2016Cityscapes}.

\textbf{Mask2former.} 
Mask2former \cite{cheng2021Mask2former} is a powerful general-purpose segmentation model in recent years with excellent performance on semantic segmentation, instance segmentation and panoramic segmentation. It uses learnbale initialization queries to represent an object, including the things class and stuff class, and updates the queries by transformer's self-attention and cross-attention. We use the pixel-decoder and transformer decoder of Mask2former as our segmentation head. Pixel-decoder performs full semantic capture on the feature map output from the backbone network to construct multi-scale features and a large resolution mask feature with rich semantic information. In the transformer decoder, the learnable query interacts with the multi-scale feature through cross-attention and further enriches the semantic information through self-attention to finally update the learnable query. Finally, the output of the category is obtained by classifying the learnable query, and the output of the mask is obtained by its dot product with the mask feature.
We choose Mask2former because it has been proven to achieve superior semantic segmentation performance compared to UperNet. However, the pre-trained weights of Mask2former that utilizes large-scale backbones are not open-source.

\subsection{Post-processing}

\textbf{CascadePSP.}  CascadePSP \cite{cheng2020cascadepsp} is a general cascade segmentation refinement model which can refine any given input segmentation predictions. It can boost the performance of segmentation models without finetuning.
Although trained on low-resolution datasets, CascadePSP can be used to produce high-quality and very high-resolution segmentation results.
CascadePSP splits the segmentation prediction by class into binary predictions, refines each binary prediction individually, and then reassembles them to obtain refined prediction results.
The success of CascadePSP on semantic segmentation datasets lead us to choose it as our post-processing method.

\textbf{SAM.} Segment Anything Model (SAM) \cite{kirillov2023segany} produces high quality arbitrary object masks from input prompts such as points or boxes, and it can be used to generate masks for all objects in an image. It has been trained on a dataset of 11 million images and 1.1 billion masks, and has strong zero-shot performance on a variety of segmentation tasks. However, SAM lacks the ability to predict semantic categories for each mask. Semantic Segment Anything (SSA) \cite{chen2023semantic} proposes a pipeline on top of SAM to predict semantic category for each mask. SSA can boost performance on close-set semantic segmentation based on any segmentation model without fintuning.
Due to the remarkable segmentation capabilities of SAM, we choose to use SAM-based SSA as another post-processing method.



\section{Experiments}

In this section, we first introduce the dataset and implementation details. Then, we provide the experiment results to validate the effectiveness backbone and segmentation heads. Finally, we report the final results on the final test set.

\subsection{Dataset Used}

\textbf{ADE20K. } Ade20k \cite{zhou2019semantic} is a comprehensive dataset that provides rich scene annotations, including object instances, object parts, and detailed semantic segmentation labels. The SceneParse150 benchmark which is constructed based on the top 150 object categories ranked by their total pixel ratios within the dataset has indeed emerged as a significant benchmark for the semantic segmentation of images. And in our experiments, we attempt to fine-tune the VSPW dataset using weights pretrained on ADE20K.


\textbf{COCO-Stuff. } COCO-Stuff \cite{caesar2018cocostuff} augments the popular COCO \cite{lin2014microsoft} with pixel-wise annotations for a rich and diverse set of 91 stuff classes. It contains 172 classes: 80 thing, 91 stuff, and 1 class unlabeled and 164k images. The 80 thing classes are the same as in COCO and the 91 stuff classes are curated by an expert annotator. Since COCO is about complex, yet natural scenes containing substantial areas of stuff, COCO-Stuff enables the exploration of rich relations between things and stuff. Therefore COCO-Stuff offers a valuable stepping stone towards complete scene understanding. So we pretrain our model on the COCO-Stuff dataset to enhance the model's scene understanding capabilities.

\textbf{VSPW. } The Video Scene Parsing in the Wild (VSPW) \cite{miao2021vspw} dataset is the first multi-scene large-scale video semantic segmentation dataset, covering more than 200 video scenes. At a frame rate of 15 f/s, this dataset is densely annotated with 3536 videos and 251633 semantically segmented frames, covering 124 semantic categories. Among these videos, 2806 videos with 198244 frames is set for training use, while other 343 videos with 24502 frames for validation and 387 videos with 28887 frames for test. Over 96\% of the video data in this dataset has a resolution between 720p and 4K. But it also provides the dataset with the resolution of 480P, that is what we use for training and testing.

\subsection{Implementation details}

\textbf{Training strategy.} We use an open source classical semantic segmentation framework MMsegmentation \cite{mmseg2020}, as the basic framework to build the model. We also employ the implementation of Mask2former from \cite{chen2022vitadapter}.
For experiments of encoder, we train the combinations of Swin-L, ConvNext-XL, InternImage-L, and InternImage-H with UperNet. Except for InternImage-H, we initialize the models with ADE20K pre-training weights. For InternImage-H, we initialize it  with both Joint 427M and ADE20K pre-training weights. Additionally, we train Mask2former combined with InternImage-H pre-trained on Joint 427M and COCO-Stuff, respectively.
Regardless of which pre-training weights we use, we adopt default training settings with different backbones for training on the semantic segmentation dataset. Due to limitation of computational resource, we adjust the training iterations and batchsize.
In addition, since the resolution of the VSPW dataset being 480*853, we use a crop size of 480*853 during training instead of using an equal width and height crop size. Code is available at \url{https://github.com/DUT-CSJ/PVUW2023-VSS-3rd}.

\textbf{Inference. }We use test time augmentation(TTA) such as stochastic flipping, multi-scale data enhancement strategies, and used model ensemble during inference. The scales for multi-scale inference are (0.5, 0.75, 1.0, 1.25, 1.5, 1.75). In the final stage of the challenge, we use CascadePSP and SAM-based SSA as post-processing methods.

\subsection{Experiment results}

\vspace{-2mm}
\begin{table}[h]
\caption{{\small{}Comparison of performance on VSPW validation set and test set1 in terms of mIoU for models trained with different backbones. All models in table use UperNet as decoder. The results in the table are obtained without using any TTA.}}\label{tab: table1}
\vspace{-3mm}
\scalebox{0.95}{
\centering{}%
\setlength{\tabcolsep}{6pt}
\begin{tabular}{cc|cc}
    \hline 
    \toprule
    \multicolumn{1}{c}{Backbone} & Pre-trained & val mIoU & test1 mIoU \\
    \midrule
    \multicolumn{1}{c}{Swin-L} & ADE20K & 56.24 & 48.09 \\
    \multicolumn{1}{c}{ConvNeXt-XL} & ADE20K & 55.50  & 47.61 \\
    \multicolumn{1}{c}{InternImage-XL} & ADE20K &  56.12  & 49.24 \\
    \multicolumn{1}{c}{InternImage-H} & ADE20K & \textbf{64.62} & \textbf{59.15} \\
    \bottomrule
    \hline
\end{tabular}
}
\vspace{-3mm}
\end{table}

\vspace{-2mm}
\begin{table}[h]
\caption{{\small{}Comparison of performance on VSPW validation set and test set1 in terms of mIoU for models trained with different decoders and pre-trained strategies. All models in table use InternImage-H as backbone. The results in the table are obtained without using any TTA.}}\label{tab: table2}
\vspace{-3mm}
\scalebox{0.89}{
\centering{}%
\setlength{\tabcolsep}{6pt}
\begin{tabular}{cc|cc}
    \hline 
    \toprule
    \multicolumn{1}{c}{Decoder} & Pre-trained & val mIoU & test1 mIoU \\
    \midrule
    \multicolumn{1}{c}{UperNet} & ADE20K & 64.62 & 59.15 \\
    \multicolumn{1}{c}{Mask2former}& Joint 427M & 66.09 & 59.69 \\
    \multicolumn{1}{c}{Mask2former}& COCO-Stuff164K & \textbf{68.03} & \textbf{62.31} \\
    \bottomrule
    \hline
\end{tabular}
}
\vspace{-3mm}
\end{table}

\vspace{-2mm}
\begin{table}[h]
\centering
\caption{{\small{}Comparison of performance on VSPW test set1 in terms of mIoU for models using TTA. * indicates employing Mask2former as decoder, \# is pre-trained on COCO-Stuff164K.}}\label{tab: table3}
\vspace{-3mm}
\scalebox{0.62}{
\centering{}%
\setlength{\tabcolsep}{6pt}
\begin{tabular}{cc|c}
    \hline 
    \toprule
    \multicolumn{1}{c}{Model} & TTA & test1 mIoU \\
    \midrule
    \multicolumn{1}{c}{Swin-L+ConvNeXt-XL+InternImage-XL} & multi-scale \& flip \& ensemble & 52.56 \\
    \multicolumn{1}{c}{InternImage-H+InternImage-H*}& ensemble & 60.12 \\
    \multicolumn{1}{c}{InternImage-H*\#+InternImage-H}& ensemble & 58.53 \\
    \multicolumn{1}{c}{InternImage-H*\#}& multi-scale \& flip & 62.60 \\
    \multicolumn{1}{c}{InternImage-H*\#}& (0.5,0.75,1.0) \& flip & \textbf{62.62} \\
    \bottomrule
    \hline
\end{tabular}
}
\vspace{-3mm}
\end{table}

\vspace{-2mm}
\begin{table}[h]
\centering
\caption{{\small{}Comparison of performance on VSPW final test set in terms of mIoU for models with different post-processing methods. All models in table use InternImage-H as backbone and Mask2former pre-trained on COCO-Stuff164K as decoder.}}\label{tab: table4}
\vspace{-3mm}
\scalebox{1.0}{
\centering{}%
\setlength{\tabcolsep}{6pt}
\begin{tabular}{c|cc}
    \hline 
    \toprule
    \multicolumn{1}{c}{Post-processing}& test mIoU \\
    \midrule
    \multicolumn{1}{c}{-}& 63.37 \\
    \multicolumn{1}{c}{CascadePSP}& \textbf{63.65} \\
    \hline
    \multicolumn{1}{c}{-}& 64.09 \\
    \multicolumn{1}{c}{SSA}& \textbf{64.11} \\
    \bottomrule
    \hline
\end{tabular}
}
\vspace{-3mm}
\end{table}

\textbf{Results of different backbones with UperNet. }Tab.~\ref{tab: table1} shows the results of different backbones with UperNet without TTA. 
We can observe that Internimage-H outperforms Swin-L, ConvNeXt-XL, and Internimage-XL significantly in terms of mIoU. Among them, Internimage-H with UperNet has a parameter count of 1.12B compared to 368M for Internimage-XL with UperNet, indicating that the increase in parameters enhances the model's performance. 

\textbf{Results of different decoders and pre-trained strategies.}
Tab.~\ref{tab: table2} shows the results of different decoders and pre-trained strategies with InternImage-H as backbone without TTA. Despite not being trained on semantic segmentation dataset, Mask2former as decoder performs better than UperNet as decoder. We also pre-train the Mask2former model on COCO-Stuff164K, which resulted in a performance improvement of 2.62 on the test set.
Tab.~\ref{tab: table1} and Tab.~\ref{tab: table2} show that larger backbone, more advanced decoder, and increased amount of pre-training data can all enhance model performance.

\textbf{Results of TTA. } Tab.~\ref{tab: table3} shows the results of TTA. We employ multi-scale flip and model ensemble as TTA approaches. 
The default scale for multi-scale inference is set to (0.5, 0.75, 1.0, 1.25, 1.5, 1.75), and the default weight for each model in the model fusion is 1. It can be observed that both multi-scale flip inference and model ensemble contribute to performance improvements. However, for the best-performing single model, InternImage-H with Mask2former pre-trained on COCO-Stuff164K, the gains from multi-scale flipped inference are minimal, and ensemble it with other models actually degrades performance. Therefore, we ultimately submit the predictions from best-performing single model with multi-scale and flip inference.

\textbf{Results of post-processing methods. }Tab.~\ref{tab: table4} shows the results of employing different post-processing methods. 
Due to time constraints, we are unable to apply post-processing to all well-performing prediction results. We use different prediction results for post-processing, where the initial predictions were obtained from model that uses InternImage-H as backbone and Mask2former pre-trained on COCO-Stuff164K as decoder. CascadePSP uses single-scale prediction as initial prediction, and SSA apply multi-scale flipped prediction at scales of (0.5, 0.75, 1.0). These initial prediction results are not the best-performing ones on the final test set. Therefore, although post-processing yield some performance improvements, it still do not surpass the best-performing results on the final test set.

\section{Conclusion}

Large models have demonstrated excellent performance across various computer vision tasks. 
During the course of the competition, we explore the performance of different backbones, decoders, and pre-training strategies in semantic segmentation. Experiments indicate that the model with Internimage-H as the backbone, Mask2former as the decoder, and pre-training using COCO-Stuff164K achieves the best results. It shows that larger backbone, more advanced decoder, and increased amount of pre-training data can all enhance model performance. We also explore the use of CascadePSP and SAM as post-processing techniques, and although they do not achieve the best results on the final test set, they are proven to enhance performance. Finally, our approach achieves 64.84\% mIoU on VSPW final test set under multi-scale flipped inference and secures the third position in the PVUW2023 VSS track.

{\small
\bibliographystyle{ieee_fullname}
\bibliography{egbib}
}

\end{document}